\documentclass[10pt,twocolumn,letterpaper]{article}

\usepackage{cvpr}              

\usepackage{subfig}
\usepackage{graphicx}
\usepackage{amsmath}
\usepackage{amssymb}
\usepackage{booktabs}
\usepackage{color}
\usepackage{algorithm}
\usepackage{algpseudocode}
\usepackage{paralist}
\usepackage{comment}
\usepackage{tabularx}
\usepackage{enumitem}
\usepackage{gensymb}
\usepackage[normalem]{ulem}
\usepackage{lipsum}

\usepackage{calc}
\usepackage[export]{adjustbox}
\usepackage{tikz}
\usetikzlibrary{positioning}
\usetikzlibrary{calc}

\newcommand\blfootnote[1]{%
  \begingroup
  \renewcommand\thefootnote{}\footnote{\hspace{-1.5\parindent}#1}%
  \addtocounter{footnote}{-1}%
  \endgroup
}

\definecolor{darkgreen}{RGB}{30,150,30}
\definecolor{darkblue}{RGB}{0,0,127}
\definecolor{darkyellow}{RGB}{171,133,0}
\definecolor{darkred}{RGB}{180,20,20}
\definecolor{darkmagenta}{RGB}{127,0,127}
\definecolor{darkcyan}{RGB}{0,127,127}
\definecolor{chromeyellow}{rgb}{1.0, 0.65, 0.0}
\definecolor{amber}{rgb}{1.0, 0.75, 0.0}

\newif\ifdrafting
\draftingfalse 

\ifdrafting
  \newcommand{\OG} [1] {\textcolor{darkgreen}{[OG: #1]}}
  \newcommand{\HS} [1] {\textcolor{darkred}{[HS: #1]}}
  \newcommand{\AB} [1] {\textcolor{darkmagenta}{[AB: #1]}}
  \newcommand{\DL} [1] {\textcolor{darkyellow}{[DL: #1]}}
  \newcommand{\TODO} [1] {{\color{darkcyan}{\bf [TODO: #1]}}}
    
\else
  \newcommand{\OG} [1] {}
  \newcommand{\HS} [1] {}
  \newcommand{\AB} [1] {}
  \newcommand{\DL} [1] {}
  \newcommand{\TODO} [1] {}
  
\fi

\definecolor{cvprblue}{rgb}{0.21,0.49,0.74}
\usepackage[pagebackref,breaklinks,colorlinks,citecolor=cvprblue]{hyperref}

\usepackage[capitalize]{cleveref}
\crefname{section}{Sec.}{Secs.}
\Crefname{section}{Section}{Sections}
\Crefname{table}{Table}{Tables}
\crefname{table}{Tab.}{Tabs.}

\title{FoVA-Depth: Field-of-View Agnostic Depth Estimation for \\Cross-Dataset Generalization}

\author{Daniel Lichy$^{\dagger,\diamond}$ \hspace{3mm}
Hang Su$^{\diamond}$ \hspace{3mm} 
Abhishek Badki$^{\diamond}$ \hspace{3mm}
Jan Kautz$^{\diamond}$ \hspace{3mm}
Orazio Gallo$^{\diamond}$\\
$^{\dagger}$University of Maryland \hspace{3mm} $^{\diamond}$NVIDIA\\
}

\begin{document}
\maketitle

\begin{abstract}

Wide field-of-view (FoV) cameras efficiently capture large portions of the scene, which makes them attractive in multiple domains, such as automotive and robotics.
For such applications, estimating depth from multiple images is a critical task, and therefore, a large amount of ground truth (GT) data is available.
Unfortunately, most of the GT data is for pinhole cameras, making it impossible to properly train depth estimation models for large-FoV cameras.
We propose the first method to train a stereo depth estimation model on the widely available pinhole data, and to generalize it to data captured with larger FoVs.
Our intuition is simple: We warp the training data to a canonical, large-FoV representation and augment it 
to allow a single network to reason about diverse types of distortions that otherwise would prevent generalization. 
We show strong generalization ability of our approach on both indoor and outdoor datasets, which was not possible with previous methods.\blfootnote{\url{https://research.nvidia.com/labs/lpr/fova-depth/}}

\end{abstract}

\def\s{\mathbf{s}}
\def\Real{\mathbb{R}}
\def\so{\mathfrak{so}}
\def\SO{\mathbb{SO}}
\def\1{\mathbf{1}}
\def\xy{\text{xy}}
\def\Z{\mathbf{Z}}
\def\d{\mathbf{d}}
\def\C{\mathcal{C}}
\def\tW{\widetilde{\W}}
\def\tD{\widetilde{\D}}
\def\td{\widetilde{\d}}
\def\tbPsi{\widetilde{\bPsi}}
\def\tP{\widetilde{P}}

\def\idx{{(i)}}

\newcommand{\SfM}{S\textit{f}M\xspace}
\newcommand{\SfC}{S\textit{f}C\xspace}
\def\SfMpp{S\textit{f}M++\xspace}

\newcommand{\centered}[1]{\begin{tabular}{l} #1 \end{tabular}}

\section{Introduction}\label{sec:intro}

Multi-view stereo (MVS), the task of estimating depth from multiple overlapping images, has applications in autonomous driving, robotics, real estate capture, and others.
Large field-of-view (FoV) images, \eg, fisheye or 360\degree~equirectangular projection (ERP) images, and the corresponding depth estimates, capture larger portions of the scene with fewer images compared with pinhole images, making them attractive for automotive and real estate applications.
The challenge, however, is the scarcity of datasets with ground truth depth for large-FoV images needed to train the depth estimation models.

Can we use the abundant small-FoV data and generalize to large-FoV fisheye and ERP data instead?

Distortion is one main challenge for generalization across FoVs.
This is because the image of a given object is distorted based on the location at which the corresponding rays intersect the image plane, and the prominence of this effect is a function of the FoV, with larger FoVs inducing larger distortion.
Intuitively, distortion makes it harder to learn generalizable image features and hinders matching across different images.
As a result, existing methods either introduce datasets for the specific cameras they target, or can only be applied to cameras for which public datasets with GT data are available.

\begin{figure}
    \includegraphics[width=\columnwidth]{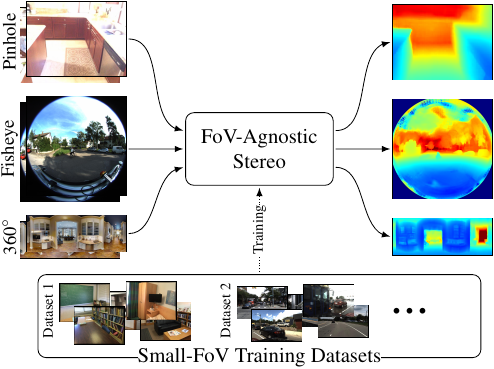}
    \caption{Our FoV-agnostic MVS model can be trained on small-FoV pinhole data and generalizes to images of various FoVs and camera models at inference time.}\label{fig:teaser}
\end{figure}

Assume we want to estimate depth for a fisheye image, but we only have a model trained on pinholes.
We could extract several rectified pinholes from the fisheye, compute depth for each, and combine the estimates back into a large-FoV depth map, as done by Rey-Area~\etal for monocular depth estimation~\cite{rey2022360monodepth}.
This requires running inference multiple times and optimizing the results together.
A more efficient solution would be to define a representation that minimizes distortion, such as a cubemap~\cite{cube_padding_salicancy,wang2020bifuse}, and train a model using only pinhole images appropriately mapped to this representation.
In this paper, we describe a surprisingly simple data augmentation strategy that enables this strategy, and show that it works even for representations that do not minimize distortion, such as ERP.

We note that most common image models, including pinhole, fisheye, and ERP are central~\cite{Ramalingam2017AUM}, \ie, they capture rays arriving at a single point, the \emph{center of projection}. 
We call these Generalized Central Cameras (GCCs). 
This property allows us to conduct \emph{extrinsic rotation augmentation} (ERA), which simulates rotating the camera about its center of projection at training time. 
ERA warps the original images to different locations of the large-FoV representation (cubemap, ERP, fisheye, \etc), forcing the network to learn to reason about distortions from only pinhole data.

\begin{figure}
    \includegraphics[width=\columnwidth]{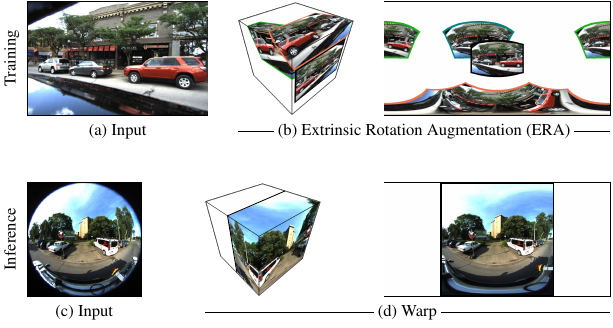}
    \caption{To estimate FoV-agnostic depth, we warp the inputs to a target representation (\eg, cubemap or ERP). We introduce Extrinsic Rotation Augmentations so that images are warped to all areas of this representation at training time (b). This forces the model trained on pinhole data to learn to reason about distortions in other types of images.}
    \label{fig:augmentation}
\end{figure}

To leverage this insight, we adapt sphere-sweeping stereo to use large-FoV representations (see Figure~\ref{fig:overview} for the case of cubemap).
Specifically, we warp the input images to a canonical representation independent of the original camera model (Figure~\ref{fig:augmentation}(b)) and train it with ERA.
This procedure works for different target representations, provided that the warped pinhole images span the whole target surface during training.
That is, ERA must cover all locations on the cubemap (including those that straddle multiple cube faces), or all areas of the ERP (including across its borders), which introduces the need for padding.
We show that properly dealing with padding is critical, and propose convolution operators for cubemap (Section~\ref{subsec:cubeconv}) and ERP (Section~\ref{subsec:circpad}) for processing cost volumes.

We demonstrate improved cross-FoV generalization of our model with respect to the state-of-the-art in both indoor and outdoor scenarios. 
For the indoor case, we train on the small-FoV dataset ScanNet~\cite{dai2017scannet} and test on 360\degree~ERP images from Matterport360~\cite{Matterport3D}. 
For outdoor, we train on the small-FoV DDAD dataset~\cite{packnetDDAD} and test on 180\degree~fisheye images in the KITTI360 dataset~\cite{Liao2021ARXIVKitti360}.
In summary, our contributions in this work include
\begin{itemize}[nosep]
    
    \item A generalized framework for MVS that works for arbitrary GCC images;
    \item The introduction of extrinsic rotation augmentations, which allow us to train on pinhole images and generalize to arbitrary GCC images, even with significantly larger FoVs;
    \item The necessary modifications to the convolution operations needed to perform MVS on cubemap and ERP.
 \end{itemize}

\section{Related Work}\label{sec:related}
We discuss prior works studying small and large-FoV stereo and MVS, and spherical data representations used for other tasks.
We will also discuss how some of these works fit into our generalized framework described in Section~\ref{sec:background}, their limitations, and how we resolve them.

\paragraph{Small-FoV MVS.}
Depth estimation from stereo and MVS pinhole cameras is one of the most widely studied topics in computer vision.
We discuss a few inspirational works here and point the reader to surveys of traditional~\cite{scharstein2002taxonomy,furukawa2015mvstutorial} and learning-based~\cite{zhu2021mvspsvdl,wang2021mvsdl} approaches.
Most modern learning-based approaches are based on the ideas proposed by GC-Net~\cite{kendall2017stereocv}, which uses a learning-based cost-volume filtering approach for stereo depth estimation.
MVSNet~\cite{yao2018mvsnet} and DeepMVS~\cite{huang2018deepmvs} extended this idea by allowing fusing information from multiple cameras.
Our framework is inspired by MVSNet and can be seen as generalizing it by adapting convolution operations to work with arbitrary GCCs.

Many works extend and improve upon the basic idea of MVSNet, which can similarly apply to our approach. 
Several methods~\cite{casmvsnet,yang2020cvpyramid,cheng2020adaptivevolume,mi2022generalized} increase efficiency and accuracy by adopting a pyramidal approach, which can be seen as applying MVSNet iteratively. 
Others improve the feature fusion stage~\cite{guo2019groupwisecorr} or adopt transformer architecture at various stages of the pipeline~\cite{ding2022transmvsnet,wang2022mvster, cao2023mvsformer}.

\paragraph{Large-FoV MVS.}

With the widespread availability of large-FoV fisheye and 360$\degree$ cameras, there has been significant interest in using them for multi-view depth estimation.
Most previous learning-based approaches take advantage of a particular geometry structure like fixed stereo~\cite{wang2019360sdnet,CubeStereoPaper} or multi-view~\cite{OmniMVS,won2020end,SweepNet,crownconv,Li_Jin2022MODE,wang2019360sdnet, Li_Jin2022MODE} rigs.

Wang~\etal~\cite{wang2019360sdnet} and Li~\etal~\cite{CubeStereoPaper} use ERP and hybrid cubemap-ERP~\cite{wang2020bifuse} representations, respectively, for the stereo setting where cameras are placed on top of each other.
MODE~\cite{Li_Jin2022MODE} shows that for any two ERP images there are extrinsic rotations to simulate the cameras being on top of each other and performs image rectification. 
Although this can be done for general two-camera configurations, they only study it for a fixed side-by-side ERP configuration.
Some multi-view works~\cite{OmniMVS,won2020end,SweepNet} do sphere sweeping from a central location between a fixed four-fisheye camera rig and perform cost-volume filtering using standard CNNs.
Komatsu~\etal~\cite{crownconv} also perform sphere sweeping from a central view, however they extract features by projecting images onto an icosahedron.
On the other hand, Li~\etal~\cite{Li2021csdnet} extract features on a spherical mesh.
These works either rely on fixed camera rigs or specialized convolution operations that are non-trivial to extend. 
A notable exception is~\cite{chiu360mvsnet}, which adapts a standard MVS architecture, CasMVSNet~\cite{casmvsnet}, by simply replacing the pinhole model with the ERP model.

These works rely on the availability of large-FoV GT depths for training.
In contrast, Lee~\etal\cite{lee2022semisup360mvs} use semi-supervised training on real images. However, their network architecture is the same as~\cite{OmniMVS}, designed for a fixed rig.
In this work, we outline a general MVS framework and a training strategy that works for arbitrary GCCs and can be trained using small-FoV datasets.

\paragraph{Spherical Data Representations.}
Several methods have studied the problem of applying neural networks to spherical data.
The most relevant to our work are those that demonstrate applying convolutional networks to cubemap~\cite{wang2020bifuse,cube_padding_salicancy} and ERP~\cite{wang2018omnidirectional,zioulis2018omnidepth} images.
To handle discontinuities while applying 2D convolutions, 2D padding operations were introduced for cubemap~\cite{wang2020bifuse,cube_padding_salicancy} and ERP~\cite{wang2018omnidirectional}.
We extend these padding operations for different stages of MVS pipelines and show that they are critical.

Orthogonal methods use ERP representations of spherical data, but deform the convolution shape on the ERP such that it is less deformed in the spherical domain~\cite{tateno2018distortion,coors2018spherenet,su2019kernel}. 
Other works proposed representing spheres and spherical convolutions with spherical harmonics~\cite{cohen2018sphericalcnns, esteves2018sphericalcnns}. 
However, both approaches require significant modifications of convolution operations and have limited support from existing libraries, making them non-trivial to extend to for MVS pipelines.
For a more detailed survey on spherical data representations, we refer the reader to \cite{gao2022parnoramicsurvey}.

\section{Preliminaries}\label{sec:background}
We introduce generalized central cameras (GCCs), state their properties, and define generalized sphere sweeping, which is then used to describe a general MVS pipeline.

\subsection{Generalized Central Camera (GCC)}\label{sec:generalized_camera}
\label{subsec:generalize_camera_model}
An image is a measurement of the light field, where the camera intrinsics and extrinsics describe how points on the camera sensor relate to physical rays. 
Mathematically, light field is a function $L: \Real^3 \times \mathbb{S}^2 \rightarrow \Real^c$, such that $L(o,\omega)$ is the $c$ color channel observed at position $o$ from direction $\omega$.
We model the generalized sensor as a 2D surface, $U$, its intrinsics as an injective function $\phi: U \rightarrow \mathbb{S}^2$, and its extrinsics as rotation and translation, $(R,t)$.
The image captured by this camera is then defined as  $I(u) = L(t, R\phi(u))$. 
The tuple $(U,\phi)$ defines the GCC. 
Similar definitions can be found in \cite{Ramalingam2017AUM,Grossberg_generalizd_cams}. 
Note that GCCs may not correspond to a physical camera, hence the term \emph{generalized}. 
A number of examples fitting the GCC definition are given in Table~\ref{tb:example_gccs}.

\begin{table}
    \centering
    \resizebox{\columnwidth}{!}{
    \begin{tabular}{c | c c}
        \hline
        GCC & $u \in U$ & $\phi(u)$ \\ 
        \hline
        Pinhole $K$ & $(u_x,u_y) \in [-1,1] \times [-1,1]$ & $K^{-1}(u_x,u_y,1)^{\intercal} / \|K^{-1}(u_x,u_y,1)^{\intercal}\|$ \\
        ERP & $ (u_x,u_y) \in [0,2\pi] \times [0,\pi]$ & $(\sin(u_y)\sin(u_x), \cos(u_y), \sin(u_y)\cos(u_x))$  \\
        Fisheye &  $(u_x,u_y) \in [-1,1] \times [-1,1]$ & various (see Supplementary) \\
        Cubemap & $u \in \mathbb{C}=\{x \in \Real^3: ||x||_{\infty}=1\}$ &  $u/\|u\|$ \\
        Sphere & $u \in \mathbb{S}^2=\{x \in \Real^3: ||x||=1\}$ &  $u$ \\ \hline
    \end{tabular}}
    \caption{GCC Examples. Note that the pinhole camera is actually a family of GCCs parameterized by intrinsics $K$.} 
    \label{tb:example_gccs}
\end{table}

\subsection{Image Warping and Extrinsic Rotations}
\label{subsec:image_warping_extrinsics_rotation}

Given an image $I$ from GCC $(U,\phi)$ and extrinsics $(R,t)$, we can warp $I$ to synthesize an image $I'$ taken with a different GCC $(U',\phi')$ and a different extrinsic rotation $R'$, but the same translation $t$, using the formula:
\begin{equation}
    I'(u) = I(\phi^{-1}(R^{-1}R'\phi'(u)).
    \label{eq:warping_between_cameras}
\end{equation}
Note that $\phi$ is surjective only when the FoV covers the full sphere, otherwise some pixels in $I'$ are not defined. 
We call surjective GCC models \textit{universal} because they capture images that allow us to synthesize the entire image of any other GCC at the same location.
Examples of universal GCCs include ERP, cubemap, and sphere.
In practice, images are discrete, and thus we need interpolation for $u$ in Equation~\ref{eq:warping_between_cameras}.

\subsection{Generalized Sweeping}
\label{subsec:generalized_sphere_sweeping}
The foundation of all sweeping-based MVS methods is warping source images onto a reference image using depth or distance hypotheses.
We focus on distance as it more naturally lends itself to FoVs larger than 180$\degree$. 
Specifically, consider two images $I^0$ and $I^i$, with GCCs $(U^0,\phi^0)$ and $(U^i,\phi^i)$\footnote{We use $I^i$ rather than $I^1$ to be consistent with the following sections.}. 
Let $(R^i,t^i)$ be the transformation that takes points in the coordinate system of camera $i$ to that of camera $0$. 
A pixel $u \in U^0$ maps to a locus in $U^i$, which we can identify by lifting $u$ to 3D using a hypothesis distance $d$, and project it back onto $U^i$.
This locus is a generalized epipolar line, which for GCCs is not necessarily straight, as we show in the Supplementary.
Formally, the reprojected point $E^i$ is given by:
\begin{equation}\label{eq:reprojected}
    E^i(u,d) = (\phi^i)^{-1} \frac{(R^i)^{-1}\left( d \phi^0(u) - t^i  \right)}{\|(R^i)^{-1}\left( d \phi^0(u) - t^i  \right)\|}.
\end{equation} 
Equation~\ref{eq:reprojected} allows us to test distance hypotheses.
That is, $\hat{d}$ is the correct depth for $u$ if $I^i(E^i(u,\hat{d}))$ and $I^0(u)$ are the image of the same 3D point, barring occlusions.
This warping is analogous to homography warping in plane-sweeping~\cite{yao2018mvsnet}. 
Finally, Equation~\ref{eq:reprojected} can easily be adapted to have the distance hypotheses depend on $u$, making this formulation suitable also for multi-stage methods~\cite{casmvsnet,mi2022generalized,chiu360mvsnet}.

\subsection{General MVS Pipeline}
\label{subsec:mvspipeline}

We can then proceed to describe a general pipeline that is shared by most sweeping-based MVS methods. 

\paragraph{Inputs.}
The inputs are a set of $N$ images from known GCCs and extrinsics. 
Assume a standard GCC is given.
We can warp all images to it using Equation~\ref{eq:warping_between_cameras}, 
yielding $N$ images $I^i$, with shared intrinsics $(U_s, \phi_s)$, and with extrinsics $(R^i,t^i)$, where $i \in \{0\dots N-1\}$. 
Rotation during warping is not strictly necessary, however, we later show it is critical for generalization (Section~\ref{subsec:era}). 
We call $I^0$ the reference image and $I^1,\dots, I^{N-1}$ the source images. 
\paragraph{Cost Volume.}

A cost volume is a data structure for facilitating pixel matching between images.
We first extract features $F^i$ for each image $I^i$. 
We then select $D$ distance hypotheses $d_j$, $j\in \{0,\dots, D-1\}$.  
For each distance hypothesis, we warp the source features to the reference view as described in Section~\ref{subsec:generalized_sphere_sweeping}. 
The warped source features for each hypothesis distance are then stacked to construct a feature volume given by:
\begin{equation}
    {\text{Vol}}^i(u,j) = F^{i} \left( E^i(u,d_j) \right).
    \label{eq:volume_construction}
\end{equation}
Interpolation is required for evaluating $F^{i}$ in Equation~\ref{eq:volume_construction}. 
Therefore the ability to allow efficient interpolation becomes critical, as we further discuss in Section~\ref{subsec:representations}. 
The reference feature volume $\text{Vol}^0(u,j)$ is created by just repeating $F^0$ along the distance hypothesis dimension. 
Next, we can fuse the features volumes from all views to form a cost volume:
\begin{equation}
    \text{CV}(u,j) = f_{\text{fusion}}( {\text{Vol}}^0(u,j), \dots, {\text{Vol}}^{N-1}(u,j) ).
\end{equation}
The intuition is that if $d_j$ is the correct distance for pixel $u$ and the features are view-invariant, then the value $(u,j)$ of all feature volumes should be the similar. 
$f_{\text{fusion}}$ is a fusion function that measures such feature similarity. 
Many MVS methods~\cite{yao2018mvsnet,casmvsnet} use variance as the fusion function, where the variance should be low across volumes when $d_j$ is the correct depth at pixel $u$.
Another choice, which we adopt in our experiments, is Group-Wise Correlation~\cite{guo2019groupwisecorr,mi2022generalized}. 

\paragraph{Distance Regression.}
Next, a 3D network is applied to the cost volume followed by a softmax operation to form a probability volume $\text{PV}$, where $\text{PV}(u,j)$ denotes the probability that the distance of $u$ is $d_j$ among all distance hypotheses. 
We can then generate the distance estimation with a weighted sum: $d^*(u) = \sum_j \text{PV}(u,j) d_{j}$. 

\paragraph{Monocular Refinement.}
Stereo matching can fail due to various reasons, \eg, flat textureless regions, occlusions, and dynamic objects.  
Furthermore, estimations are often output at reduced resolutions due to efficiency limitations. 
A 2D network can take the initial estimation, and optionally the reference image, and output a refined final estimation.

\section{Method}\label{sec:method}

\begin{figure*}
    \centering
    \includegraphics[]{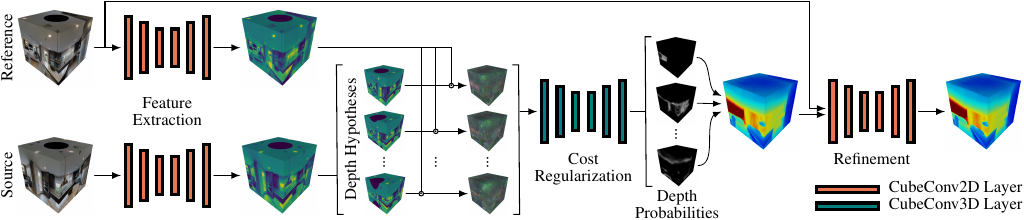}
    \caption{Our MVS pipeline. Here we only show the architecture for cubemap, but the same pipeline can be used for ERP by simply switching the convolution operations.}\label{fig:overview}
\end{figure*}

Our goal is to train an end-to-end MVS network that can operate on images taken by any GCCs. 
We use readily available small-FoV datasets for training, and design our model to generalize to large-FoV datasets at inference time. 
To accomplish this, we propose warping the input images to a canonical representation independent of their original representations.
Although many choices exist for the canonical representation, we develop our approach with ERP (Section~\ref{subsec:circpad}) and cubemap (Section~\ref{subsec:cubeconv}) due to their desirable properties outlined in Section~\ref{subsec:representations}.
We also introduce a data augmentation strategy at training time that is critical to generalization (Section~\ref{subsec:era}). 
After warping, we can apply the general MVS pipeline described in Section~\ref{subsec:mvspipeline} adapted to the canonical representation (Figure~\ref{fig:overview}).

\subsection{Extrinsic Rotation Augmentation}
\label{subsec:era}
The first step of our approach is to warp each image to the canonical representation using Equation~\ref{eq:warping_between_cameras}, where $(U,\phi)$ is the original GCC of the image and $(U',\phi')=(U_s,\phi_s)$.  
Since we train on small-FoV data, if we only warp the images with $R'=R$, the resulting images may not exhibit the types of distortion seen in wide-FoV images, \eg, on edges and corners of cubemap or near the top and bottom of ERP. 
This hinders the ability of the network to generalize.
To mitigate this issue, we propose \emph{extrinsic rotation augmentation} (ERA) as a type of data augmentation during training.
In a nutshell, we warp the image to the canonical representation using a random rotation, $R'$, so as to span all the regions of the canonical representation, forcing the network to learn about distortion (Figure~\ref{fig:augmentation}).

\subsection{Canonical Representations}
\label{subsec:representations}
Any GCC can be used as a canonical representation.
However, we identify three properties that an ideal canonical representation for the MVS pipeline should possess:

\begin{enumerate}[nosep]
    \item \label{prop_phi_onto} It should be universal (Section~\ref{sec:background}), so we can represent images with FoVs up to full $360\degree$;
    \item \label{prop_transfer_learning} It should be compatible with existing deep networks to allow the use of strong architectures and enable transfer learning;
    \item \label{prop_interpolation} It should allow for interpolation at arbitrary locations efficiently for fast construction of cost volumes.
\end{enumerate}
Property~\ref{prop_phi_onto} requires our representation to be bijective to the sphere, and therefore any GCC can be warped to it without cropping any pixels.
An intuitive option is some meshing of the sphere, \eg, an icosphere~\cite{crownconv}.
Instead, we turn to the ERP and cubemap representations.
ERP and cubemap cover $360\degree$ FoVs, thus satisfying Property~\ref{prop_phi_onto}.
Moreover, standard CNN architectures can be trivially adapted to operate efficiently on ERP~\cite{zioulis2018omnidepth,wang2020bifuse} and cubemap~\cite{cube_padding_salicancy,wang2020bifuse}. 
Finally, these two GCCs lend themselves to efficient GPU-accelerated interpolation.

In principle, low distortion is another desirable property for a GCC.
This would favor cubemap, which incur minimal distortion due to the small FoV of each face. 
However, empirically we found that the ERA strategy allows our models to perform well even for ERP despite the larger distortion. 

\subsection{Equirectangular Projection (ERP)}
\label{subsec:circpad}
ERP is widely used for 360\degree~panoramas in the industry, because it is intuitive to visualize and easy to work with. 
Since ERP data are represented with standard 2D images, we can simply adopt standard 2D CNN operations and architectures without any modifications. 
However, without proper care for padding, the quality of the results degrades measurably as the errors along the sides increase.

\paragraph{Circular Convolution.} ERP captures the full 360$\degree$~FoV and we can leverage this property to mitigate the issue above.
Because of their circular nature, the left and right boundaries of ERP images connect to form a continuous horizon, which allows us to pad at the image boundaries with content from the opposite sides.
Note that to really achieve a continuous horizon, circular padding needs to be done at every convolutional layer, not just at the input layer. 
While it is possible to preemptively pad the input up to the width of the receptive field of the CNN, this can incur substantial overhead.

Instead, we use circular padding for each layer in isolation, and only pad the necessary amount.
We call the convolution layer that uses this padding strategy Circular Convolution, or CircConv (Figure~\ref{fig:padding}(a)). 
Formally, given an input feature map $F$ of height $H$ and width $W$, we define the horizontally padded feature map as:
\begin{equation}
    \hat{F}[i,j] = F[i, (j-P)~\text{mod}~W],
\end{equation}
where $i,j \in [0,H) \times [0,W+2P)$. We use zero padding for the vertical sides only, though strategies for the top and bottom are also possible~\cite{wang2018omnidirectional}. 

\paragraph{CircConv3D.} For the cost-regularization network, we need the 3D analog of CircConv.
To achieve that, we simply split the volume into a list of ERP images, one for each depth hypothesis, and apply CircConv to them individually.
Last, we can concatenate the list to form a volume and apply a standard 3D convolution.

\subsection{Cubemap Representation}
\label{subsec:cubeconv}

\begin{figure}
    \centering
    \includegraphics[width=\columnwidth]{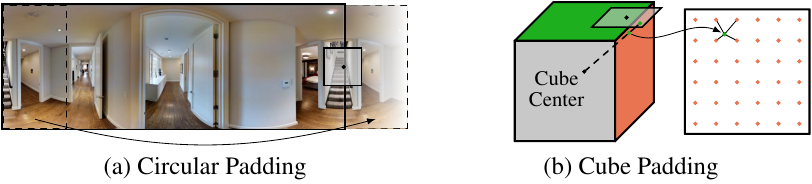}
    \caption{(a) We pad a side of the ERP by replicating pixel values from the opposite side. (b) We pad the green cube face with the interpolated value of the green point projected on to the orange face. Transparent squares indicate the convolution filters.
    }\label{fig:padding}
\end{figure}

Standard CNN architectures can work for the cubemap data in the MVS pipeline by simply operating separately on each face. 
In this way, though, the relationship between faces is lost, causing excess artifacts along the cube edges and overall more restrictive receptive fields. 
Similar to CircConv in the case of ERP, we need a mechanism to bring back the continuity along boundaries. 

\paragraph{Cube Convolution.} We resolve this using a similar strategy to the ERP case:
we define a convolution operator on the cube that wraps the filters around the cube edges, in a similar manner that CircConv wraps filters around the sides of the ERP. 
Just like for ERP, this is implemented using padding followed by standard convolution.
In particular, we adapt the spherical padding of Wang~\etal~\cite{wang2020bifuse}, but replace bilinear interpolation with nearest neighbor interpolation for an efficient implementation.
This is equivalent to the cube padding of Cheng~\etal~\cite{cube_padding_salicancy} for the case of $3 \times 3$ convolution, which is used in all layers in our architecture except the first one of the feature extractor.

Concretely, assume a $c$-channel cubemap image $I_{\text{cube}}$ taken by a camera, $(\mathbb{C}, \phi_{\text{cube}})$, is represented as an array $F$ of shape $(6,c,W,W)$. 
Let us consider a $(2P+1) \times (2P+1)$ CubeConv to the $0$-th face, $F_0 = F[0,:,i,j]$.
To pad $F_0$ with padding of size $P$, we first extend the sampling locations beyond the cube face to $i,j \in \left[-P, W+P\right) \times \left[-P, W+P\right)$. 
Then we project the extended sample locations back onto cube and interpolate their values to produce the padded $0$-th face, denoted as $\hat{F}_0$:
\begin{align}
    \hat{F}_0 = I_{\text{cube}}(\phi_{\text{cube}}^{-1}(\phi_0(\frac{2i}{W}-1,\frac{2j}{W}-1))), 
\end{align}
where $\phi_0$ is the intrinsics of a pinhole camera corresponding to the $0$-th cube face. 
Figure~\ref{fig:padding}(b) shows this process.
$I_{\text{cube}}(\cdot)$ is evaluated at subpixel locations with interpolation.
With the padding in place, each face can be filtered individually and together generates an output of shape $(6,c',W, W)$.

\paragraph{CubeConv3D.} The 3D case is handled by treating the volume as a list of cubes analogous to CircConv3D.

\paragraph{Face Culling.}
In practice, when training on small-FoV images, some cube faces will be empty. 
We skip these faces to reduce time and memory consumption. 
These skipped faces may still be queried for cube padding but can simply be treated as all zeros.

\subsection{Reciprocal Tangent Sampling}
Constructing and processing 3D cost volumes is expensive, so the number of distance hypotheses is limited, and it is critical to sample the distance range efficiently. 
We therefore propose using \emph{reciprocal tangent sampling}, defined as 
\begin{equation}
    \{d_{j}\} = f_{\text{RT}}(\mathcal{U}(f_{\text{RT}}^{-1}(d_{\min}), f_{\text{RT}}^{-1}(d_{\max}),D)),
\end{equation}
where $f_{\text{RT}}(x) = \frac{2}{\pi \tan(\frac{\pi}{2} x)}$, and $\mathcal{U}(v_{\min},v_{\max},D)$ is a function that uniformly samples $D$ points between $v_{\min}$ and $v_{\max}$.
The intuition is that the angular disparity between two images is roughly related to distance by the inverse tangent function.
We show in Section~\ref{subsec:ablation} that this sampling strategy works better than the commonly used inverse distance sampling for unbounded scenes in the datasets we evaluate.
We discuss this sampling strategy and the intuition behind it in greater detail in the Supplementary.

\section{Evaluation and Results}\label{sec:results}

\subsection{Datasets}
Our primary goal is to train our network on small-FoV datasets and evaluate on large-FoV datasets. 
For the indoor scenario, we train on ScanNet~\cite{dai2017scannet} (small-FoV) and test on Matterport360~\cite{rey2022360monodepth} (large-FoV).
For the outdoor scenario, we train on DDAD~\cite{packnetDDAD} (small-FoV) and test on KITTI-360~\cite{Liao2021ARXIVKitti360} (large-FoV).
Due to the unavailability of GT depths aligned with the fisheyes in KITTI-360, and the presence of dynamic objects in multi-view images selected from different time instances, we only test generalizability for outdoor scenarios qualitatively.

\noindent\textbf{ScanNet} 
consists of 94,212 stereo pairs from 1,201 indoor scenes.
The images are all captured with pinhole cameras with FoVs $<$ 60$\degree$. 
We use the same data split as \cite{Kusupati_2020_CVPR}.

\noindent\textbf{Matterport360} 
consists of 9,684 RGB-D ERP images from 90 building-scale scenes. We only use the 18 test scenes in the official split.
It was originally designed for 360$\degree$ single-image depth estimation, so we choose any two images within 2 meters of each other as stereo pairs.
The images capture the entire 360$\degree$ view, except for the regions around the poles.
We choose Matterport360 over existing large-FoV stereo datasets such as Deep360~\cite{Li_Jin2022MODE} and Stanford2D3D~\cite{stanford2d3d} in order to analyze the large-baseline settings, where distortion across views differs significantly.
The stereo image pairs in Deep360 and Stanford2D3D are nearby and have a fixed orientation relative to each other.

\noindent\textbf{DDAD}
consists of 200 driving sequences captured from 6 pinhole cameras.
We use the first 150 scenes for training and scenes 150--159 for validation.
We use every image as a reference image and select the frame forward in time that has a camera displacement closest to 1 meter as the source view.
For three-view experiments, we additionally select the frame backward in time with a camera displacement closest to 1 meter as the second source view.

\noindent\textbf{KITTI-360}
consists of 11 driving sequences captured using two 180$\degree$ fisheye cameras on the sides of the vehicle and a front-facing perspective stereo camera.
We build image pairs or triplets in the same manner as in DDAD.

\subsection{Implementation Details}
All models are implemented in Pytorch. We use NVDiffRast~\cite{Laine2020diffrast} for fast interpolation of the cubemap. 
We apply $L1$ loss on log depth for the estimated depth maps both before and after monocular refinement. 
We use 48 distance hypotheses and a cube size of $6 \times 256\times 256$ or an ERP size of $384 \times 1024$. 

\paragraph{Networks.}
For the feature extractor, we use the first three blocks of ResNet34~\cite{he2016deep} with all convolution layers replaced with CircConv or CubeConv layers, depending on the chosen canonical representation. 
We then use transposed convolution layers to upsample all feature maps to 1/4 input resolution and concatenate them to form the final image feature. 
For the cost regularization network, we use the MVSNet architecture~\cite{yao2018mvsnet} with all 3D convolution layers replaced with CircConv3D or CubeConv3D layers. 
Finally, we use the MiDaS~\cite{ranftl2020towards} architecture for the refinement network, again with all convolutions replaced with CircConv or CubeConv.  
More implementation details can be found in the Supplementary. 

\paragraph{Metrics.}
We use standard metrics widely used for depth estimation~\cite{eigen2014depth}. 
The metrics include AbsRel (absolute relative error), RMSE (root mean square error), and percentage measures $\max( \frac{\hat{y}}{y}, \frac{y}{\hat{y}}) < \delta$ for $\delta=1.25,1.25^2,1.25^3$.

\subsection{Baseline Evaluations}

\paragraph{Quantitative Comparison for Indoor Scenes.} 
Since we are the first to tackle the problem of training on small-FoV stereo images and generalizing to large-FoV images, there are no off-the-shelf baselines. 
Therefore, we propose two baselines. The first is the state-of-the-art rectified ERP stereo method, MODE~\cite{Li_Jin2022MODE}, which we retrain using ScanNet. 
Since MODE is only designed to operate on rectified ERP images, we warp the ScanNet images to rectified ERP images using Equation~\ref{sec:background}. 
More details of this procedure can be found in the Supplementary. 
The second is the state-of-the-art 360 MVS method, 360MVSNet~\cite{chiu360mvsnet}, which uses ERP as its underlying representation. 
Since code is not available, we reimplement it ourselves and train on ScanNet. We also improve 360MVSNet with an upgraded ResNet feature extractor, 360MVSNet-ResNet, and our ERA, 360MVSNet-ResNet-ERA. 
We compare both of these baselines to our method using the ERP representation and our method using the cubemap representation, both with and without monocular refinement. 
The results can be found in Table~\ref{tb:matterport_results}. 

Surprisingly, we observe similar quality between our method with the cubemap and the ERP, despite the larger distortion of the ERP. 
This demonstrates the power of the ERA in learning to deal with distortion even from pinhole images. 
Once we upgrade 360MVSNet's feature extractor to the same ResNet as our models and apply our proposed ERA we obtain similar performance to our ERA model without refinement. 
This is not surprising since this is basically the same as our ERP model with the additional cascade processing. 
Moreover, we observe that monocular refinement actually performs better than the cascade strategy on the evaluated datasets.
Note that, due to the required image rectification, training MODE with ERA is not trivial.

\begin{table}
    \resizebox{\columnwidth}{!}{
    \begin{tabular}{ c | c c c c c c}
     \hline
    Method & AbsRel $\downarrow$ & SqRel $\downarrow$ & RMSE $\downarrow$ & $\delta1$ $\uparrow$ & $\delta2$  $\uparrow$ & $\delta3$ $\uparrow$  \\ \hline
    MODE~\cite{Li_Jin2022MODE}                & 0.459    & 0.873   & 1.292 & 0.23   & 0.494  & 0.763  \\  
    360MVSNet-FCN~\cite{chiu360mvsnet}        & 0.477    & 1.256   & 1.191 & 0.579  & 0.745  & 0.839  \\  
    360MVSNet-ResNet     & 0.367    & 0.821   & 0.994 & 0.654  & 0.794  & 0.869  \\  
    360MVSNet-ResNet-ERA & 0.236    & 0.39    & 0.779 & 0.724  & 0.846  & 0.907 \\ \hline
    Ours Cube            & 0.232    & 0.445   & 0.763 & 0.745  & 0.857  & 0.911  \\  
    Ours ERP             & 0.236    & 0.465   & 0.813 & 0.736  & 0.853  & 0.911  \\  
    Ours Cube+R          & 0.186    & 0.289   & \textbf{0.665} & 0.78   & 0.879  & 0.925  \\  
    Ours ERP+R           & \textbf{0.170}    & \textbf{0.249}   & 0.668 & \textbf{0.791}  & \textbf{0.895}  & \textbf{0.944}  \\  \hline
    \end{tabular}}
    \caption{Comparison of methods on Matterport. +R = with monocular refinement}
    \label{tb:matterport_results}
\end{table}

\begin{figure*}
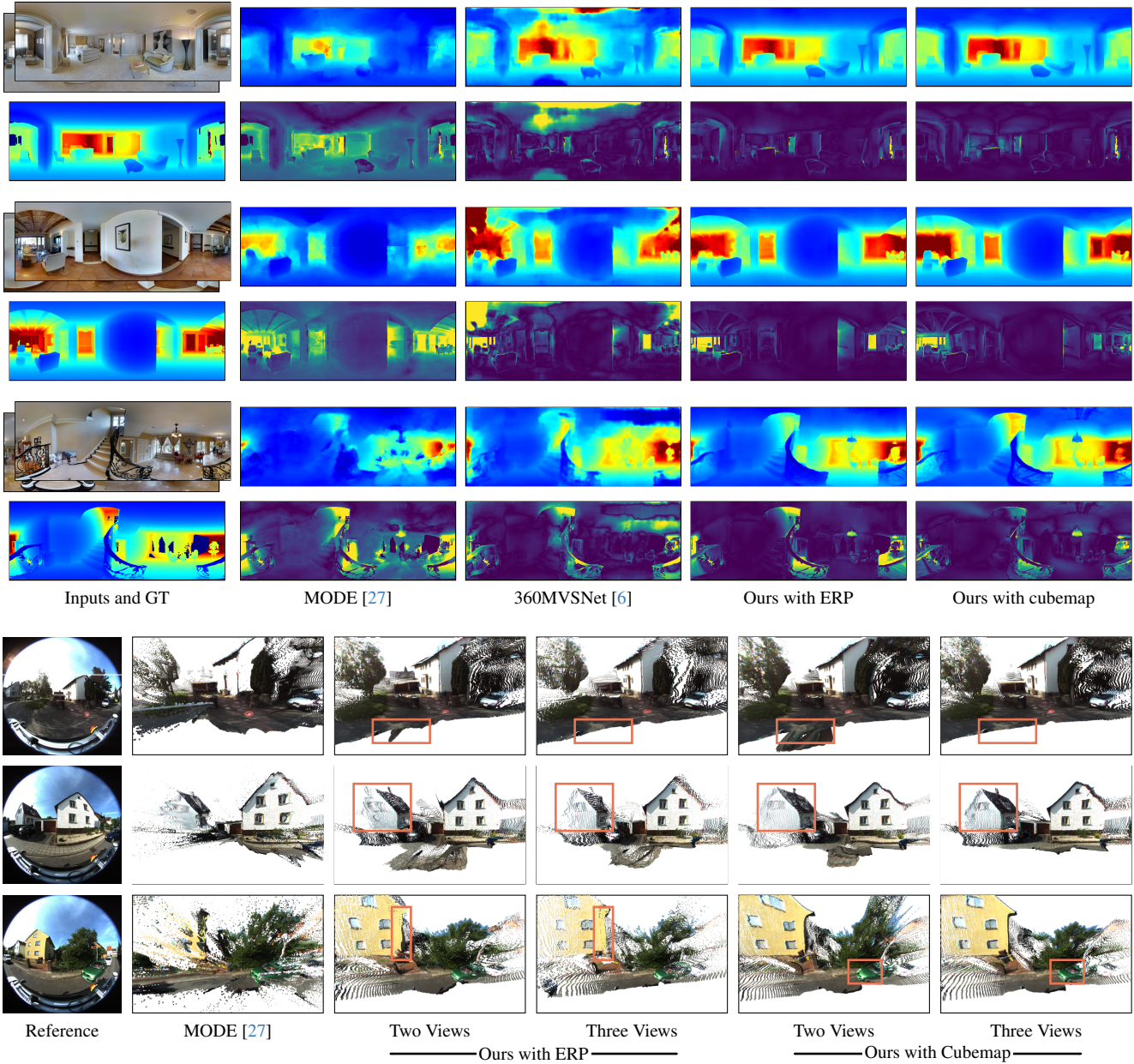

    \centering
    \input{figures/results_matterport/result_matterport}
    \input{figures/results_kitti/result_kitti}
    \caption{Generalization results of our approach on Matterport360 (top) and KITTI-360 (bottom). For indoor scenes, our approach trained with ERA for both ERP and cubemap representations outperform competing approaches~\cite{Li_Jin2022MODE,chiu360mvsnet}. For outdoor scenes, our approach generalizes better than MODE~\cite{Li_Jin2022MODE} trained only on large-FoV synthetic data. Our approach can naturally use additional views. Our 3-view stereo shows better reconstructions (see the highlighted regions) for both ERP and cubemap representations.}\label{fig:results}
\end{figure*}

\paragraph{Qualitative Comparison for Outdoor Scenes.}
To demonstrate the advantages of training on real small-FoV data versus synthetic large-FoV data, we finetune our ScanNet-trained model with images from DDAD for up to 100 epochs. We compare it to the MODE model pre-trained by the authors on their large-FoV synthetic driving dataset, Deep360~\cite{Li_Jin2022MODE}. 
We evaluate both models qualitatively on the large-FoV KITTI360 dataset. 
Qualitative results are shown in Figure~\ref{fig:results}.

\subsection{Ablation Studies}
\label{subsec:ablation}
\paragraph{Role of Extrinsic Rotation Augmentation.}
We show that ERA is essential for generalization for both ERP and cubemap. 
Table~\ref{tb:extrinsics_rotation_results} shows the comparison of models with and without the augmentation. 
ERA helps all models generalize both with and without monocular refinement.

\begin{table}
    \resizebox{\columnwidth}{!}{
    \begin{tabular}{ c | c c c c c}
     \hline
    Method & Ours Cube & Ours Cube+R & Ours ERP & Ours ERP+R & 360MVSNet-ResNet \\ \hline
    w/ ERA  & \textbf{0.232} & \textbf{0.186} & \textbf{0.236} & \textbf{0.17}  & \textbf{0.235} \\ 
    w/o ERA & 0.413 & 0.375 & 0.417 & 0.376 & 0.367 \\ \hline
    \end{tabular}}
    \caption{Comparison of models with and without extrinsic rotation augmentation. AbsRel numbers are reported here.}
    \label{tb:extrinsics_rotation_results}
\end{table}

\paragraph{Role of CubeConv.}
To investigate the benefit of CubeConv, we train a version of our model with standard convolution layers and zero padding. 
There is a drop in performance both with and without monocular refinement as shown in Table~\ref{tb:other_ablations}. 
Qualitatively, we see large seams form around the edges of the cubemap (see Supplementary). 

\paragraph{Role of Pre-training.} 
One advantage of CubeConv is that it is compatible with standard network architectures. 
This enables transfer learning, which allows training high-performing models with limited training data.  
To demonstrate this, we train a variant of our model with random initialization of the feature extractor and monocular refinement network instead of using ImageNet pre-trained weights. 
We again see a large drop in performance both with and without monocular refinement (Table~\ref{tb:other_ablations}).

\paragraph{Role of Reciprocal Tangent Sampling.}
We compare our cubemap model trained with reciprocal tangent sampling versus inverse distance sampling. 
We observe a significant benefit of adopting reciprocal tangent sampling (Table~\ref{tb:other_ablations}). 

\begin{table}
    \centering
    \small
    \begin{tabular}{  c | c c}
     \hline
    Method & Ours Cube & Ours Cube+R   \\ \hline
    Full model                    & 0.232 & 0.186 \\ 
    w/o pre-training               & 0.279 & 0.243 \\ 
    w/o cube padding              & 0.381 & 0.466 \\ 
    w/o recip.\ tangent sampling  & 0.262 & 0.220  \\ \hline
    \end{tabular}
    \caption{Ablations for the cubemap model. AbsRel numbers are reported here.}
    \label{tb:other_ablations}
\end{table}

\section{Conclusions}\label{sec:conclusions}
In this work, we introduce a multi-view stereo framework for Generalized Central Cameras that can be trained on small-FoV pinhole data and generalize to any cameras, including ones with large-FoV.
We show that a surprisingly simple data augmentation strategy, extrinsic rotation augmentation, is the key to enabling this generalization capability.
We adapt our MVS framework for ERP and cubemap representations by introducing efficient padding operations for convolutions for different stages of an MVS pipeline. 
We see utility for this model in automotive and real-estate applications. 
Furthermore, the method can be easily extended to leverage improvements proposed for standard pinhole sweeping methods, \eg, multi-scale and self-supervised techniques.

{\small
\bibliographystyle{ieeenat_fullname}
\bibliography{egbib}
}

\clearpage
\setcounter{page}{1}

\section{Supplementary}
\subsection{Epipolar Geometry}
\begin{figure*}
    \includegraphics{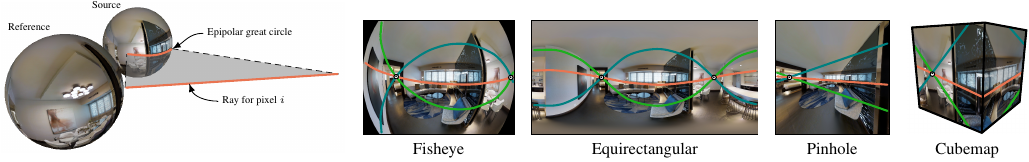}
    \caption{Epipolar Geometry for GCCs. The \emph{epipolar plane}, in gray, contains the spherical camera centers and the query ray. We call the great circle where this plane intersects the source sphere the \emph{epipolar great circle}. On the right, we show how the epipolar great circle manifests as different types of \emph{epipolar curves} in image space.}\label{fig:epipolar_geometry}
\end{figure*}

Figure~\ref{fig:epipolar_geometry} shows how epipolar geometry plays out for different GCCs. In particular, there is an \emph{epipolar plane} that contains the camera centers and the query ray. The intersection of this plane with the source sphere makes an \emph{epipolar great circle} on the source sphere corresponding to the possible matches of the query ray. The path projects, via $\phi^{-1}$ to different \emph{epipolar curves} depending on the camera model.

Note that there are two epipoles corresponding to the two intersections of the line passing through the camera centers with the reference camera. In the pinhole case, due to the FoV being strictly less than 180$^\circ$, at most one epipole is observed. In the rectified case neither epipole is observed, but in some sense one epipole is infinitely to the left and the other is infinitely to the right.

\subsection{Interpolation}
In this section, we elaborate on the necessity of property 3 and explain why general samplings of the sphere can be problematic. When we create the cost-volume, we have to warp the source images to the reference images. This process requires interpolating the source images. To perform interpolation, we need to conduct a k-nearest-neighbor search in the canonical representation. For example, for a planar image like the Equirectangular Projection (ERP), finding the nearest neighbor is essentially free. We just need to take the floor and ceiling of each pixel coordinate to obtain the four nearest neighbors for interpolation. The process for a cube is slightly more complex, since we first have to determine which face of the cube the point is on, i.e., find the coordinate with a value equal to $\pm1$. Then, it reduces to the flat plane case.

If we have more exotic samples of the sphere, we must perform a full nearest neighbor search of the points, which is prohibitively expensive for network training (at least with existing implementations). Komatsu \etal actually conduct this nearest-neighbors search. They can do so because they use a fixed camera rig, meaning the relative pose between the source and reference image is fixed. This allows them to perform the nearest-neighbors search once for the entire training period~\cite{crownconv}.

\subsection{Reciprocal Tanget Sampling}

In this section we motive our reciprocal tangent sampling strategy for generating hypotheses. 
First, consider the rectified stereo case. The disparity is related to depth by $\text{disparity} \sim 1/\text{depth}$. This implies uniformly sampling inverse depth hypotheses corresponds to sampling equidistant points in the source image. We would like to do something similar for spherical images \ie sample distance hypotheses that correspond to equidistant samples on the sphere. To accomplish this, we need to establish the relation between the distance and the sampling locations on the source sphere \ie the angular disparity.  

\begin{figure}[htb]
    \centering
        \includegraphics[width=0.4\textwidth,keepaspectratio]{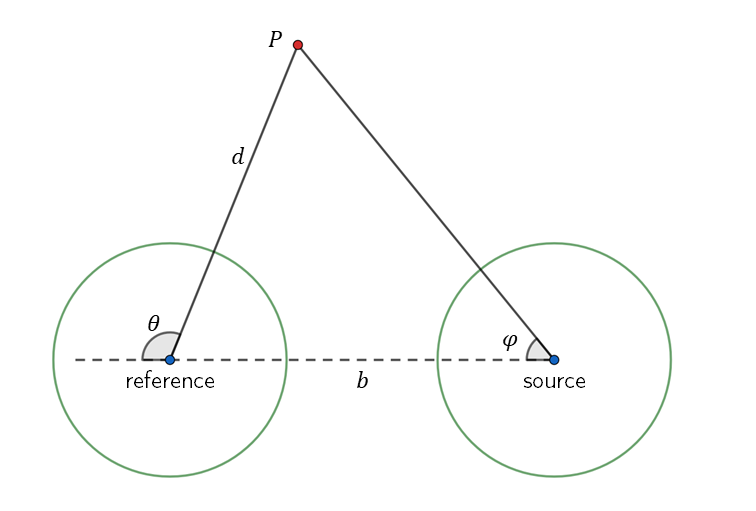} 
    \caption{Schematic drawing of the epipolar plane, gray plane in Figure~\ref{fig:epipolar_geometry}, and the angles that relate distance, $d$, to angular disparity $\alpha = \theta-\phi$. Given baseline $b$.}
    \label{fig:angular_disparity}
\end{figure}

Consider the schematic drawing of an epipolar plane for two spherical cameras shown in Figure \ref{fig:angular_disparity}. Call the vector going from the source to the reference camera the baseline vector, with length $b$.  Consider a query ray extending from the reference camera center that hits a 3D point $P$ a distance $d$ away. This ray makes an angle $\theta$ with the baseline vector. Define the angle $\phi$ as the angle between the baseline vector and the ray from the source camera center to $P$. Given $\theta$, $d$ and $\phi$ are related, though conventionally, rather than writing this relation explicitly, we write the relation between $d$ and the angular disparity, $\alpha$, defined as $\alpha = \theta-\phi$. This relation is given by:

\begin{equation}
    d(\alpha, \theta) = b \left( \frac{\sin \theta}{\tan \alpha} - \cos \theta \right)
    \label{eq:angular_disparity}
\end{equation}

Where $\theta \in [0,180]$ and $\alpha \in [0,\theta)$ \cite{wang2019360sdnet}. Note that unlike the rectified pinhole stereo case, the relation between distance and disparity depends on the reference pixel that determines $\theta$. For two cameras one can choose different hypotheses per pixel, but for more than two, this is a problem because there are different $\theta$s and $\alpha$s associated with each source camera. However, from Equation \ref{eq:angular_disparity} we observe the rough relation between disparity and distance follows a shifted and scaled inverse tangent function. This is the motivation for reciprocal tangent sampling.

Qualitatively, reciprocal tangent sampling samples points less densely near to the camera than inverse distance sampling. This fit our intuition that inverse distance sampling was wasting samples very close to the camera. An additional benefit of reciprocal tangent sampling is that it can handle arbitrarily large depths with finite samples, similarly to inverse distance and as opposed to linear sampling. This enables us to handle both indoor and outdoor scenes without any changes to the sampling strategy. 

We do not claim this sampling is optimal in all cases, just that it worked well for both indoor and outdoor scenes we studied.

\subsection{ERP Rectification for MODE}
The input to MODE \cite{Li_Jin2022MODE} must be two ERP images rectified such that the cameras are directly ``on top of each other'' that is the camera coordinate up axis (y-axis) is parallel to the cameras' translation, and both cameras have the same extrinsic rotation. The idea is to apply Equation 1 to simulate this configuration. Concretely, our inputs are two images with GCCs $(U^i,\phi^i)$ and extrinsics $(R^i,t^i)$ with $i \in {0,1}$. We write $R^i$ as the concatenation of three column vectors $R^i = \left[r^i_0, r^i_1, r^i_2 \right]$. Then mathematically to be ``on top of each other'' means $r^0_1 = r^1_1 = \frac{t^1-t^0}{||t^1-t^0||}$ and $R^0=R^1$. 

\begin{figure}[htb]
    \centering
    \begin{tabular}{cc}
        \includegraphics[width=0.22\textwidth,keepaspectratio]{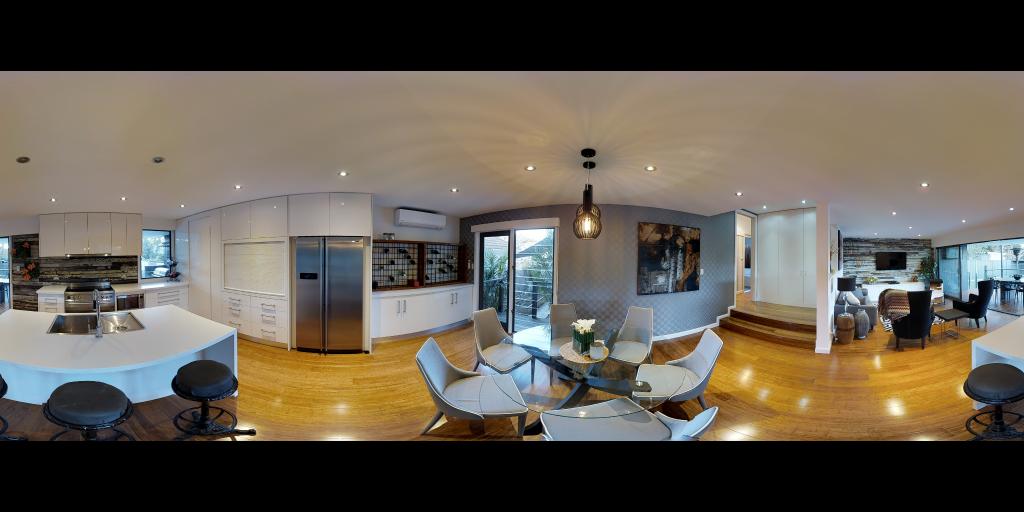} & \includegraphics[width=0.22\textwidth,keepaspectratio]{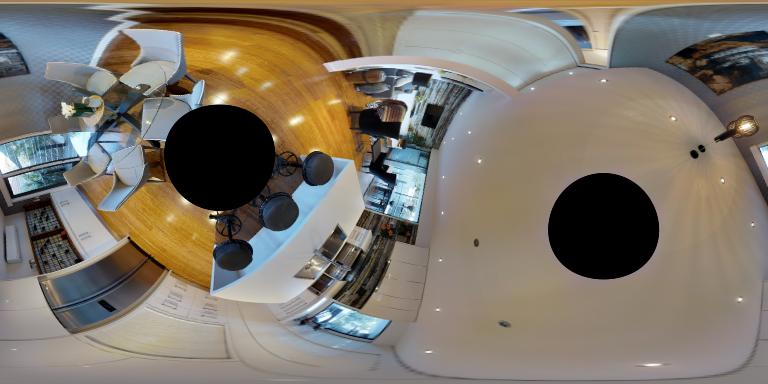} \\
        \includegraphics[width=0.22\textwidth,keepaspectratio]{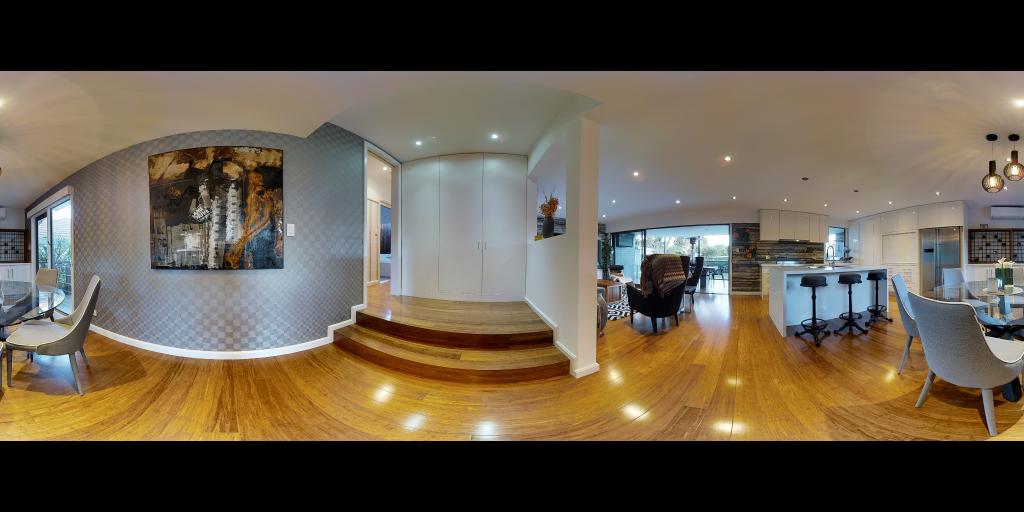} & \includegraphics[width=0.22\textwidth,keepaspectratio]{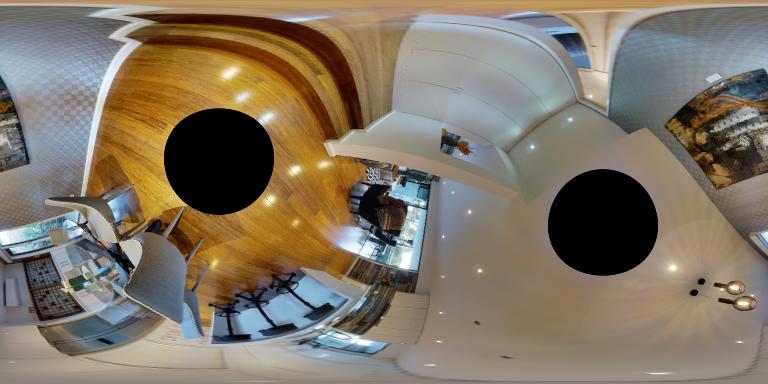} \\
    \end{tabular}
    \caption{Original ERP images (left) and rectified images for MODE (right)} \label{fig:mode_rectification}
\end{figure}

To rectify the images define $\hat{r}_1 = \frac{t^1-t^0}{||t^1-t^0||}$. Now we can choose arbitrarily $\hat{r}_2$ any vector orthogonal to $\hat{r}_1$. Define $\hat{r}_0 = \hat{r}_1 \times \hat{r}_2$, where $\times$ is the cross product of vectors. Let $\hat{R} = \left[ \hat{r}_0, \hat{r}_1, \hat{r}_2 \right]$. 
Now we warp each image using Equation \ref{eq:warping_between_cameras_supp} with $R' = \hat{R}$, $(U', \phi')$ the ERP GCC, $(U,\phi) = (U^i, \phi^i)$ and $R = R^i$:
\begin{equation}
    I'(u) = I(\phi^{-1}(R^{-1}R'\phi'(u)).
    \label{eq:warping_between_cameras_supp}
\end{equation}
These warped images are now rectified.
An example of two ERP images before and after rectification are shown in Figure~\ref{fig:mode_rectification}.

\subsection{Fisheye $\phi$}
There are many models of fisheye cameras. Usually these models model the projection function \ie $\phi^{-1}$. The model used for KITTI360 has 7 parameters. $f_1,f_2,u_0,v_0,x_i,k_1,k_2$ and is given by $\phi^{-1}(x,y,z) = (u'',v'')$ 
\begin{align*}
    (u, v) & = \left(\frac{x}{z + x_i}, \frac{y}{z + x_i}\right) \\
    r^2 & = u^2 + v^2 \\
    (u', v') & = (1 + k_1 r^2 + k_2 r^4) (u, v) \\
    (u'', v'') & = (f_1 u' + u_0, f_2 v' + v_0)
\end{align*}
where $(x,y,z) \in \mathbb{S}^2$. To compute $\phi$ we use iterative undistortion from $\phi^{-1}$~\cite{schoenberger2016mvs,schoenberger2016sfm}. Other models of fisheye cameras can be found in ~\cite{schoenberger2016mvs,schoenberger2016sfm}.

\subsection{CubePadding Seams}
\begin{figure*}[htb]
    \centering
    \begin{tabular}{ccc}
        \includegraphics[width=0.3\textwidth,keepaspectratio]{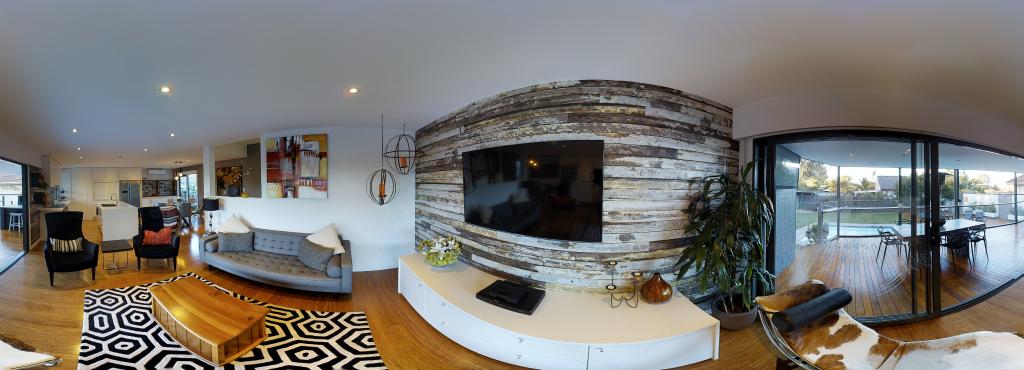} & \includegraphics[width=0.3\textwidth,keepaspectratio]{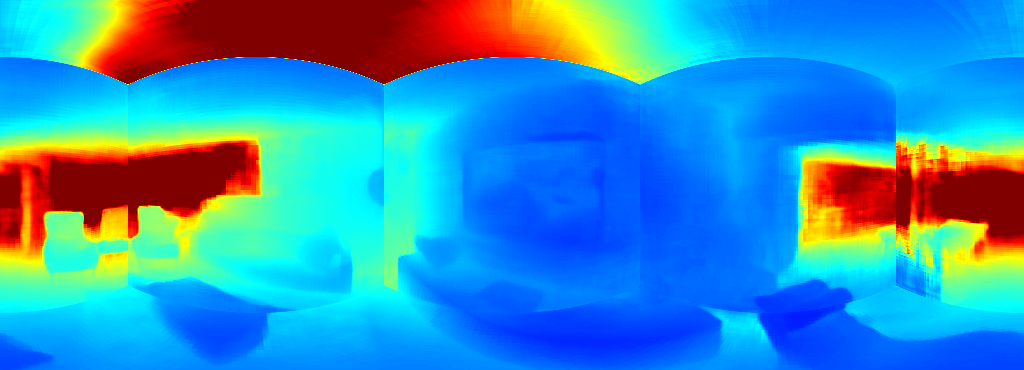} & \includegraphics[width=0.3\textwidth,keepaspectratio]{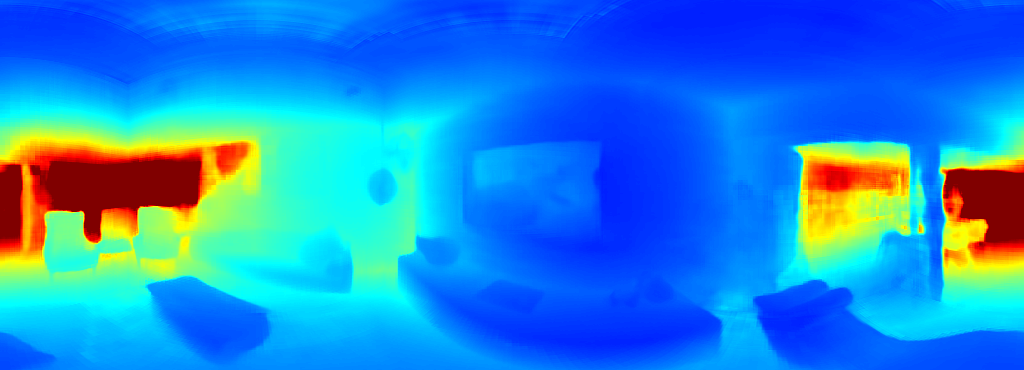}\\
        \includegraphics[width=0.3\textwidth,keepaspectratio]{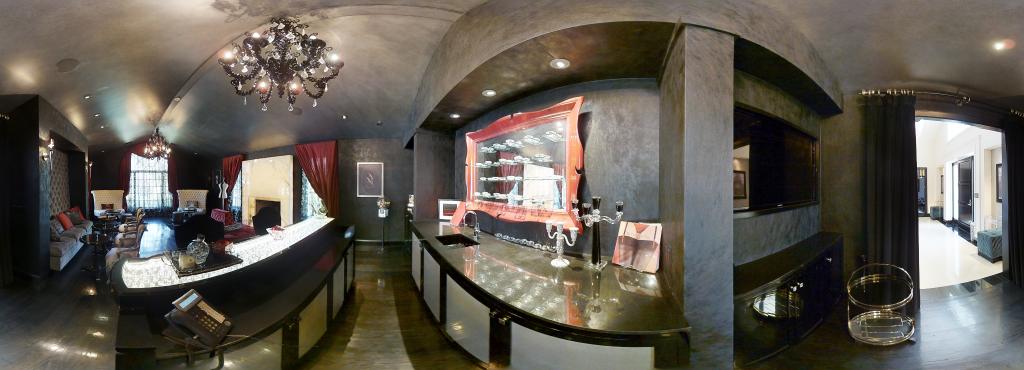} & \includegraphics[width=0.3\textwidth,keepaspectratio]{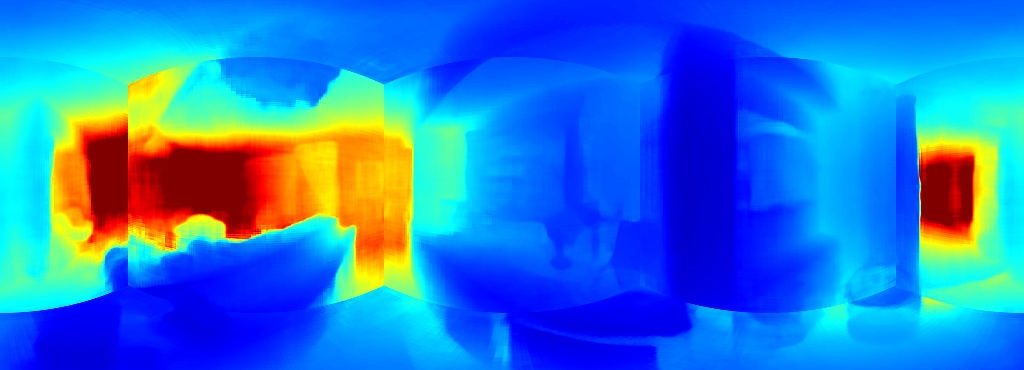} & \includegraphics[width=0.3\textwidth,keepaspectratio]{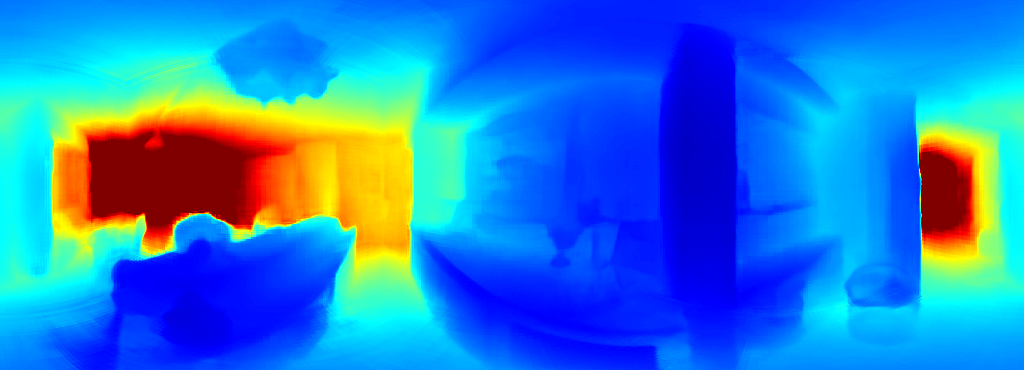}\\
    \end{tabular}
    \caption{Qualitative comparison of our cubemap model with and without cube padding.}
    \label{fig:cube_padding_qual}
\end{figure*}

Figure \ref{fig:cube_padding_qual} shows our cube model with and without cube padding. In particular, we observe strong seams on the edges of the cube faces of the model without padding. In the first row we see a large error on the textureless ceiling. This is because context from surrounding cube faces is needed to deduce the depth of the top face, but without cube padding the network has no context from surrounding faces.

\begin{figure*}[htb]
    \centering
\begin{verbatim}
    from torchvision.models.feature_extraction import create_feature_extractor
    from torchvision.models import resnet34
    
    m = torchvision.models.resnet34()
    return_nodes = {
        'layer1': 'layer1',
        'layer2': 'layer2',
        'layer3': 'layer3'
    }
    net = create_feature_extractor(m, return_nodes=return_nodes)
    features = net(image)
    f1, f2, f3 = features['layer1'], features['layer2'], features['layer3']
\end{verbatim}
    \caption{Definition of ``first three layers'' of ResNet34}
\label{fig:Resnet34_layers}
\end{figure*}

\subsection{Implementation Details}

\subsubsection{Our Models}
\paragraph{Feature Extractor} For the architecture of the feature extractor, we use the first three layers of Torchvision’s implementation of ResNet34 to extract three feature maps f1, f2, and f3 at resolutions $1/4$, $1/8$, and $1/16$ of the input resolution with  $64$, $128$, and  $256$ channels, respectively. See Figure \ref{fig:Resnet34_layers} for more details. We pass f1 though a transposed conv layer with stride 1 and 32 features, f2 though a transposed conv layer with stride 2 and 32 features, and f3 through a transposed conv layer with stride 4 and 64 features. We then concatenate the outputs to form a single feature map at $1/4$ input resolution with $128$ channels.

All convolutions are replaced with either CubeConv or CircConv depending on the respective model.

\paragraph{Cost Regularization} The cost regularization network is based on the network from MVSNet~\cite{yao2018mvsnet}. Besides replacing all convs with CubeConv3d or CircConv3d we replace all transposed convs with nearest-neighbor upsampling followed by
a conv with the same number of features as the original transposed conv. We do this to avoid the need for output padding in transposed convs in order to upsample by an exact multiple of 2.

\subsubsection{360MVSNet}
For a fair comparison, we used the same reciprocal tangent sampling to select initial distance hypotheses as we used for our method when training 360MVSNet. For subsequent stages we used the uncertainty aware sampling proposed by the authors. We considered two feature extractors: The FCN-model, which comes from casMVSNet~\cite{casmvsnet} that 360MVSNet is based on and an upgraded one based on ResNet34. The multiscale features are f1, f2, f3 as in Figure \ref{fig:Resnet34_layers}. Each is passed though a stride 1 conv layer to reduce it to 8, 16, and 32 channels respectively.

\end{document}